\documentclass[twoside,11pt]{article}

\usepackage{jmlr2e}
\usepackage{amsmath}

\ShortHeadings{Predicting Transfer to the Pediatric Intensive Care Unit with Ensemble Boosting}{Rubin, Potes, Xu-Wilson, Dong, Rahman, Nguyen \& Moromisato}
\firstpageno{1}

\begin{document}

\title{An Ensemble Boosting Model for Predicting Transfer to the Pediatric Intensive Care Unit}

\author{\name Jonathan Rubin, Cristhian Potes, Minnan Xu-Wilson,\\ 
       Junzi Dong \& Asif Rahman\\
       \email Jonathan.Rubin, Cristhian.Potes, Minnan.Xu, Junzi.Dong, Asif.Rahman@philips.com\\
       \addr Acute Care Solutions (ACS)\\
       Philips Research North America\\
       Cambridge, MA, United States
       \AND
       \name Hiep Nguyen, M.D. \& David Moromisato, M.D.\\
       \email Hiep.Nguyen, David.Moromisato@bannerhealth.com\\
       \addr Cardon Children's Medical Center\\
       Banner Health System\\
       Mesa, AZ, United States} 
       
\maketitle

\begin{abstract}
Our work focuses on the problem of predicting the transfer of pediatric patients from the general ward of a hospital to the pediatric intensive care unit. Using data collected over 5.5 years from the electronic health records of two medical facilities, we develop classifiers based on \emph{adaptive boosting} and \emph{gradient tree boosting}. We further combine these learned classifiers into an ensemble model and compare its performance to a modified pediatric early warning score (PEWS) baseline that relies on expert defined guidelines. To gauge model generalizability, we perform an inter-facility evaluation where we train our algorithm on data from one facility and perform evaluation on a hidden test dataset from a separate facility. We show that improvements are witnessed over the PEWS baseline in accuracy (\emph{0.77} vs. \emph{0.69}), sensitivity (\emph{0.80} vs. \emph{0.68}), specificity (\emph{0.74} vs. \emph{0.70}) and AUROC (\emph{0.85} vs. \emph{0.73}).
\end{abstract}

\section{Introduction}

Approximately 1\% - 3\% of pediatric patients admitted to the general ward of a hospital will be transferred to the pediatric intensive care unit (PICU) due to a deterioration in health \citep{tucker2009prospective}. Many guideline-based early warning score (EWS) systems that monitor a patient's state of health have been proposed to address this problem \citep{subbe2001validation,williams2012national,duncan2006pediatric}, as have data-driven approaches that rely on machine learned classifiers \citep{churpek2014multicenter,churpek2016multicenter}. However, the majority of these systems have focused on adult populations, with less focus on pediatric patients where it is known that vital sign measurements, such as heart rate and respiration rate differ markedly in young children compared with adolescents and adults. Moreover, existing EWS systems aimed at young populations, such as Pediatric Early Warning Score (PEWS) \citep{monaghan2005detecting,parshuram2009development}, rely on manual spot check observations made by nursing staff, such as the capillary nail refill test, which means input into the system is subjective. An automated method that detects early deterioration in pediatric patients using physiologic vital sign information offers several advantages:

\begin{enumerate}
\item It ensures that patients that are in danger of deteriorating receive timely care and attention, thereby minimizing or avoiding harm to the patient due to the occurrence of a significant adverse event.
\item It aids decision making regarding transfer of pediatric patients to an increased level of care and the allocation of hospital resources required for such a transfer.
\item It does not rely on manually recorded information that is prone to subjective bias, such as capillary refill in the PEWS system.
\item It allows an objective measurement of patient deterioration, which can reinforce the intuition of hospital care staff and act as further evidence when decisions about level of care are required to be made by a team of medical staff.
\end{enumerate}

In this work, we present the development of an automated early deterioration algorithm for pediatric populations within a hospital's general ward. Our models accept a patient's age as input, as well as physiologic vital sign measurements. This information is used to make a prediction about the likelihood of the patient transferring from the general ward to the PICU.

We compare two approaches based on ensemble boosting for creating transfer prediction models. The first relies on an \emph{adaptive boosting} algorithm \citep{schapire1999improved,conroy2016dynamic} that employs single level decision trees as its base classifier. The adaptive boosting procedure is altered to consider a patient's age for learning risk thresholds. The second approach constructs an ensemble of CART models (classification and regression trees) using \emph{extreme gradient boosting} \citep{DBLP:journals/corr/ChenG16}. Finally, we combine the predictions of both the adaptive boosting and gradient boosting models into an ensemble and evaluate its performance. We compare the results of our \emph{boosting}-based classifiers to a version of the Bedside PEWS Scoring System \citep{parshuram2009development}, which was modified based on available input data.

\section{Cohort}

Data was collected over a 5.5 year period from the electronic health records of two medical centers: Banner Thunderbird Medical Center and Banner Desert Medical Center. Encounters that occurred in the pediatric general ward(s) and pediatric intensive care unit were included in the datasets. Encounters where transfer occurred from the general ward to the PICU where determined using location and time stamp information from the electronic health record. All patients between the ages of 1 month and less than 20 years were included in the dataset. The study was approved by the Institutional Review Board (IRB) of Banner Health (Mesa, AZ, USA).

Table \ref{tab:demo} summarizes the details regarding the number of unique patients and encounters, as well as patient demographic information. The values in Table \ref{tab:demo} confirm that, for both facilities, there is a large imbalance between the number of pediatric encounters that resulted in transfer to the PICU compared to those that did not.

\begin{table}[h]
  \centering 
  \caption{Patient encounters and demographic information per hospital facility} 
  \begin{tabular}{|l|c|c|c|c|}
  	\hline
    	& \multicolumn{2}{|c|}{\textbf{Desert}} & \multicolumn{2}{|c|}{\textbf{Thunderbird}}\\
	\hline
	& \textbf{Transferred} & \textbf{Non-transferred} & \textbf{Transferred} & \textbf{Non-transferred}\\
	\hline
	\textbf{Patients}					& 305 (3.0\%)			& 9982 \: (97.0\%) 			& 	98\:\, (1.9\%)	& 	5042 (98.1\%)\\
	\textbf{Encounters}				& 330 (2.6\%)			& 12536 (97.4\%)			& 	102 (1.7\%)	& 	6005 (98.3\%)\\
	\hline
	\textbf{Average age}		& 5.4 $\pm$ 5.7	& 6.1 $\pm$ 5.8 	& 	5.5 $\pm$ 6.1	& 	6.3 $\pm$ 6.0\\
	\hline
	\textbf{Gender}					& 				& 	 			& 				& 		\\
	\textbf{ -- Female}				& 130 (42.6\%)			& 4491 (45.0\%)			& 	35 (35.7\%)			& 	2292 (45.5\%)\\
	\textbf{ -- Male}					& 174 (57.1\%)			& 5273 (52.8\%)			& 	59 (60.2\%)			& 	2622 (52.0\%)\\
	\textbf{ -- Missing}				& 1 \: (0.3\%)			& 218 (2.2\%)				& 	4 (4.1\%)				& 	128 (2.5\%)\\	
	\hline
  \end{tabular}
  \label{tab:demo} 
\end{table}

\section{Feature selection, data preprocessing and splitting}

\subsection{Feature selection}

We wished to construct a system that given a snapshot of objective inputs, our models could make a determination about the likelihood of a patient being transferred to the PICU. The following features were selected to be used as inputs into the prediction model: 1.~Heart Rate (HR); 2.~O2 Saturation (O2); 3.~Respiratory Rate (RR); 4.~Temperature (Temp); 5.~Diastolic Blood Pressure (dBP); 6.~Systolic Blood Pressure (sBP); 7.~Patient Age; 8.~Pulse Pressure ($sBP - dBP$); 9.~Approximate Mean Arterial Pressure ($\frac{2}{3}dBP + \frac{1}{3}sBP$); and 10.~Shock Index ($\frac{HR}{sBP}$).

The features listed above include direct vital sign measurements, age of the patient and three measurements derived from vital sign inputs. Laboratory values were also considered as input, as they have been included as features in adult deterioration indicators \citep{rothman2013development,churpek2014multicenter}. However, the extra stress induced in pediatric populations by performing blood draws meant that these inputs were generally collected less often and would likely be less available in practice, hence they were excluded as features. Spot check measurements such as Capillary Refill and Skin Color, originally included within PEWS systems, were excluded from the analysis due to their subjective nature.

\subsection{Data preprocessing}

For each encounter that resulted in transfer to the PICU, feature values were retrieved from the electronic health record. Feature vectors were populated from clinical event measurements that occurred at least \emph{two} hours preceding the time of transfer and at most \emph{eight} hours preceding transfer. 
The value used for each feature was the final clinical measurement recorded within the observation window, hence, each instance captured a snapshot of deterioration. Each instance that resulted in transfer was matched by a corresponding encounter that did not result in transfer. For non-transfer instances, a random \emph{six} hour observation window was selected and a snapshot of feature inputs consisting of the last recorded value in the observation window was used. In the case where no measurement was recorded for an input value within the six hour observation window that feature's value was recorded as missing.

\subsection{Data splitting}

\subsubsection{Training/Cross-validation}
\label{sec:cv}

Eighty percent of data from the Desert facility was used as training data.  10-fold cross validation was used to split this training data into separate folds. Choice of which hyperparameters to use for our models was based on maximizing the average cross-validation score over all 10 folds. Area under the receiver operating characteristic (AUROC) was used as the metric for optimization.

\subsubsection{Testing}

Twenty percent of data from the Desert facility was set aside as held-out test data. Stratified sampling was used to ensure an even class distribution.\\

\noindent{One hundred percent of data from the Thunderbird facility was set aside as a separate held-out test-set, i.e. no encounter from the Thunderbird facility was used in model training/cross-validation. This decision was made to ensure that the final results obtained on the test-set accurately reflected generalizability between individual facilities. For the Desert dataset, it was further ensured that no patient who had any encounters in the training/cross-validation sets was included in the test set.}

\section{PICU Transfer Prediction Algorithm}

We compared two variants of boosting algorithms \citep{freund1995desicion} for distinguishing between encounters that resulted in transfer to the PICU versus those that did not. Both algorithms were required to gracefully deal with missing feature values, as our dataset consisted of instances where certain vital sign information was missing and future deployment of such a system would require effective handling of missing information.

We wish to learn a model, $F_m(x) = y$, by recursively constructing baseline (``weak'') classifiers, $h(x)$, fit to a specified loss function, $L(y, F(x))$. Beginning with an initial model $F_0(x)$, the final model, $F_m(x)$, is defined recursively by combing the predictions of the previous model, $F_{m-1}(x)$, with $h(x)$.

\begin{equation}
F_m(x) = F_{m-1}(x) + \alpha h(x), \quad m \ge 1
\end{equation}

\noindent{where}, $\alpha$ is a scaling factor and $m$ is the total number of baseline classifiers to fit.

\subsection{Adaptive Boosting}

We first train an adaptive boosting classifier that seeks to add baseline classifiers, $h(x)$, that will minimize an exponential loss function:

\begin{equation}
L(y, F(x)) =  \sum_{i=1}^n w_i \cdot e^{-y^{(i)} \cdot \alpha h(x^{(i)})}
\end{equation}

\noindent{where}, $n$, is the number of instances, $w_i$ is the current weight distribution over instances and $x^{(i)}, y^{(i)}$ refers to the $i^{th}$ training data instance and label, respectively. A variation of adaptive boosting known as AdaBoost-abstain \citep{schapire1999improved,conroy2016dynamic} was employed to handle missing feature values. AdaBoost-abstain allows each baseline classifier to abstain from voting if its dependent feature is missing. We use the class of 1-dimensional decision stumps as base classifiers, where each classifier votes by comparing one feature, $x_j$, in the data to a threshold, $\tau$ and produces a classification output of +1 (positive), -1 (negative), or 0 (abstains, when the $j^{th}$ feature is missing, $x_j = \phi$).

\begin{equation}
h(x_j; \tau) = \left\{\begin{array}{rl} +1\text{, } & x_j \geq \tau \\ -1\text{, } & x_j < \tau \\ 0\text{, } & x_j = \phi \\ \end{array}\right.
\end{equation}

\noindent{To} account for age-dependent clinical features, a modification is made to the algorithm where decision thresholds are computed based on a number of predefined age groups beginning with 3 month intervals for patients less than 1 year of age ([0 -- 3 months], [3 -- 6 months], \ldots, [9 -- 12 months]) and followed by 1.5 year intervals up to age 20 ([1 -- 2.5 years] $\ldots$ [18.5 -- 20 years])). Including the patient's age in each weak classifier allows us to learn age-dependent risk thresholds and compensate for variability in the normal range of feature values over age groups. The reader is referred to \citep{conroy2016dynamic} for further algorithmic details regarding the AdaBoost-abstain procedure. The final adaptive boosting model consisted of $m=100$ decision stump base classifiers.

\subsection{Gradient Boosting}

We trained a separate classifier using gradient boosting, where $h(x)$ are recursively constructed $pseudo$-residual models, fit to the gradient of a specified loss function evaluated at $F_{m}(x)$.

\begin{equation}
h_m(x) = y - \nabla L(y, F_{m}(x))
\label{eqn:h}
\end{equation}

\noindent{In} particular, we use the XGBoost variation \citep{DBLP:journals/corr/ChenG16} of gradient tree boosting \citep{friedman2001greedy}, where CART models (classification and regression trees) are used as base classifiers and a \emph{regularized learning objective} is minimized.

\begin{equation}
\mathcal{L}(F(x)) = L(y, F(x)) + \sum_{k=1}^K \Omega(h_k(x))
\label{eqn:loss}
\end{equation}

\begin{equation}
\Omega(h_k(x)) = \gamma T + \frac{1}{2}\lambda\sum_{j=1}^T{w_j^2}
\label{eqn:omega}
\end{equation}

\noindent{Equation} (\ref{eqn:loss}) adds a regularization component, $\Omega(h_k(x))$, to the loss function that controls the model complexity of learned trees, $h_k(x)$, where $K$ is the total number of trees. In Equation (\ref{eqn:omega}), $T$ refers to the number of leaves in the tree, $\gamma$ and $\lambda$ are regularization parameters and $w$ are the leaf scores within the CART model. In our models, binary logistic loss was specified as the loss function to optimize. Age was included as a standard feature within the model allowing trees to make splits directly based on age information. Each tree also incorporated branching information based on whether feature values were missing or not, thereby directly handling missing information without requiring imputation. A random search \citep{bergstra2012random} was used to select hyperparameters such as $\gamma$ and $\lambda$, node splitting weights, number of feature columns to sample and maximum tree depth. The set of hyperparameters that optimized the cross-validation AUROC score (section \ref{sec:cv}) were selected to create a final classifier. The final gradient boosted model consisted of $m=16$ trees with a maximum depth of 3.

\subsection{Ensemble}

Finally, a simple ensemble model was created from the two boosting-based classifiers, where the average of each of the boosting models' prediction probabilities was used to make a final prediction.

\section{Modified Bedside PEWS Baseline}

We compare our algorithm's performance to the Bedside PEWS System \citep{parshuram2009development}. Bedside PEWS calculates a range of sub-scores for inputs such as heart rate, systolic blood pressure, capillary refill, respiratory rate, respiratory effort, oxygen saturation and oxygen therapy. For some inputs (heart rate, systolic blood pressure and respiratory rate) sub-scores are affected by a patient's age group. Experts defined appropriate cut-off points for each scoring item during the Bedside PEWS systems creation and validation \citep{parshuram2009development}.

Bedside PEWS was selected over other pediatric early warning score systems for comparison, as it was validated on predicting unplanned admissions to the PICU \citep{robson2013comparison}. Furthermore, the inputs of the bedside PEWS system are similar to the inputs required by our algorithm, however, some modifications were necessary. In particular, capillary refill and respiratory effort were removed as input, as they were not available within our datasets.

Given these modifications it was necessary to compute an appropriate PEWS cut-off score in which instances that match or exceed the score are classified as transfer and instances that result in a score less than the cut off score are classified as no transfer. To do so, we computed PEWS scores on our training data for all available inputs, using the expert defined scoring system from \citet{parshuram2009development}. We then evaluated each PEWS score as a potential threshold and determined the optimal cut-off score that balanced sensitivity and specificity on the training dataset. An optimal cut off score of 2 was found for the modified PEWS system. This cut off score was used to predict transfer on the test sets.

\section{Results}

\begin{table}[h]
  \centering 
  \caption{Final results for Desert facility test dataset} 
  \begin{tabular}{lcccc}
  	\hline
    	& Accuracy 		& Sensitivity & Specificity & AUROC\\
	\hline
	\hline
	PEWS			& \emph{0.74}		& \emph{0.73}		& {\bf\emph{0.75}}	& \emph{0.75}\\
	Gradient Boosting	& \emph{0.77}		& \emph{0.78} 		& {\bf \emph{0.75}} 	& 	\emph{0.83}\\
	Adaptive Boosting	& \emph{0.69}		& \emph{0.78} 		& \emph{0.60}		&	\emph{0.81}\\
	Ensemble			& {\bf \emph{0.78}}	& {\bf \emph{0.83}}	& \emph{0.73}		& 	{\bf \emph{0.84}}\\
	\hline
  \end{tabular}
  \label{tab:desert} 
\end{table}

\begin{table}[h]
  \centering 
  \caption{Final results for Thunderbird facility test dataset} 
  \begin{tabular}{lcccc}
	\hline
    	& Accuracy 		& Sensitivity & Specificity & AUROC\\
	\hline
	\hline
	PEWS			& \emph{0.69}		& \emph{0.68}		& \emph{0.70}		& \emph{0.73}\\
	Gradient Boosting	& \emph{0.76}		& \emph{0.73}		& {\bf \emph{0.78}}	& \emph{0.83}\\
	Adaptive Boosting	& \emph{0.74}		& {\bf \emph{0.84}}	& \emph{0.63}		& {\bf \emph{0.85}}\\
	Ensemble			& {\bf \emph{0.77}}	& \emph{0.80}		& \emph{0.74}		& {\bf \emph{0.85}}\\	
	\hline
  \end{tabular}
  \label{tab:thunderbird} 
\end{table}

Tables \ref{tab:desert} and \ref{tab:thunderbird} present accuracy, sensitivity, specificity and AUROC results for the Desert and Thunderbird held-out test sets, respectively. Recall that 20\% of encounters from the Desert facility were held-out as test data and 100\% of encounters from the Thunderbird facility were used as test data. For both test sets, the class prevalence is balanced, i.e. there are an equal number of instances that resulted in transfer compared to those that did not. As such, an uninformed classifier that simply predicted no transfer would achieve an accuracy of 50\%.

For accuracy, sensitivity and specificity a default threshold of 0.5 probability was used to make a prediction of transfer for each of the \emph{gradient boosting}, \emph{adaptive boosting} and \emph{ensemble} classifiers. Further tuning of this threshold on the training set could potentially allow for improved results. Tables \ref{tab:desert} and \ref{tab:thunderbird} show that the Ensemble achieves the greatest accuracy on both test datasets, \emph{0.78} and \emph{0.77} respectively. Further, for the Desert dataset a slight improvement in AUROC is witnessed using the ensemble, \emph{0.84}, whereas for the Thunderbird test set the ensemble achieved the same AUROC performance as the adaptive boosting model. In all evaluation categories one of the boosting models always outperforms the modified bedside PEWS baseline, except for specificity in Table \ref{tab:desert}, where PEWS achieves the same greatest specificity as the gradient boosting model. In general, the \emph{boosting} algorithms' results between the two facilities are not markedly different, indicating that the models constructed were able to generalize between facilities.

\begin{figure}[t]
  \centering 
  \includegraphics[width=1.0\columnwidth]{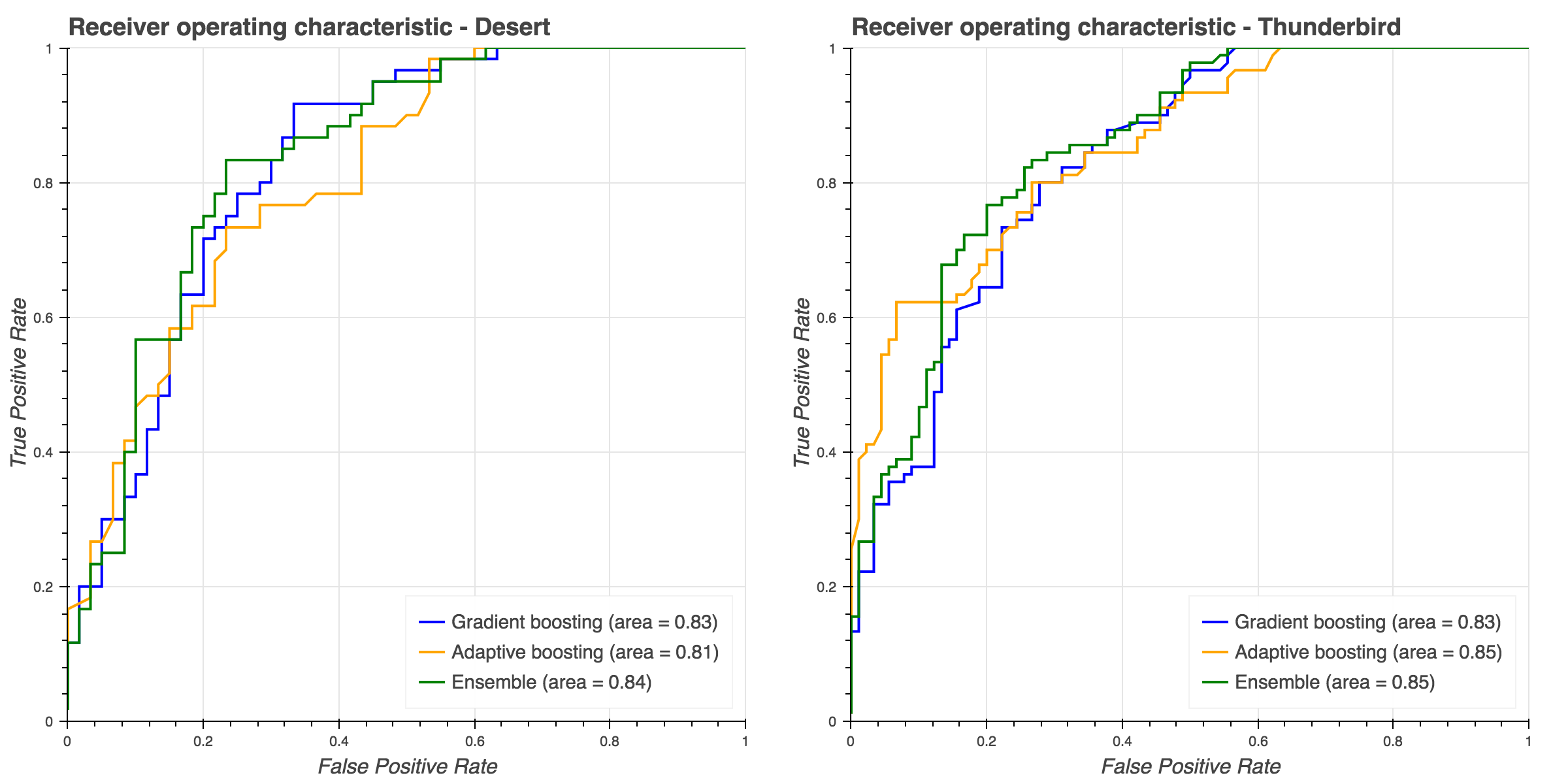} 
  \caption{Receiver operating characteristic for Desert facility (left) and Thunderbird facility (right)}
  \label{fig:roc_desert_thunderbird} 
\end{figure}

Figure \ref{fig:roc_desert_thunderbird}, shows the receiver operating characteristic for the Desert (left) and Thunderbird (right) test sets. It can be seen that when considering a tradeoff between true positive and false positive rate, the Ensemble (green line) would achieve slightly better performance for a sensible false positive range selection, e.g.~approx.~20\% false positives.

%

\begin{figure}[ht]
  \centering 
  \includegraphics[width=1.0\columnwidth]{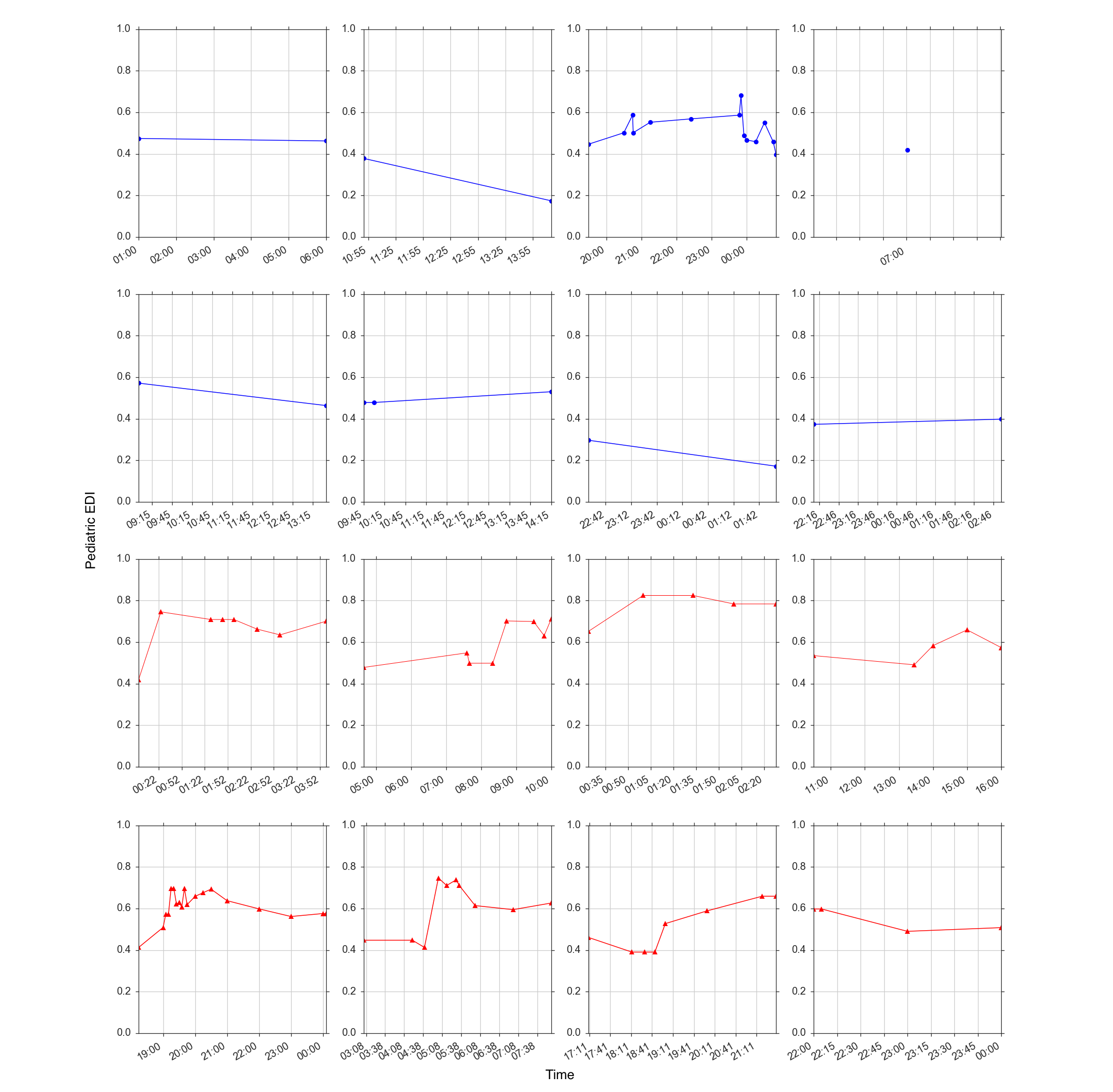} 
  \caption{Pediatric early deterioration indicator values output over six hour observation windows for eight patients that were transferred (bottom) and eight patients that were not transferred (top) to the PICU.}
  \label{fig:ts_all} 
\end{figure}

Finally, in Figure \ref{fig:ts_all} we present 16 time series plots --- the first eight patients in the test set that did not result in transfer are displayed in the above two rows and the first eight patients that did result in transfer are displayed in the two rows below. Each plot shows a six hour observation window where the ensemble model made a prediction about the likelihood of transfer whenever a new clinical measurement was recorded. For the patients that were transferred, transfer occurred two hours later within each plot.

It can be seen that the transfer prediction values are greater overall for the eight patients that were transferred, compared to those that were not transferred. Also, noticeable is the fact that patients who were transferred were more closely monitored by clinical staff, resulting in a greater number of vital sign measurements within each six hour window. In the top half of Figure \ref{fig:ts_all}, there is only one case where probability of transfer exceeds 0.6 (top row, third chart from the left). Given the number of recordings made for this patient, it does seem that deterioration was initially suspected by the care team. The probability predictions decrease towards the end of the observation window, making it seem likely that this patient stabilized and transfer was eventually not required. 

\section{Discussion and Related Work}

Our work has focused on the problem of predicting transfer for pediatric populations from the hospital general ward to the PICU. Specifically, using six vital sign values as input, as well as age and three derived measures, we constructed two boosting-based classifiers. From these classifiers a simple ensemble was created based on prediction averaging. We compared the performance of all boosting-based classifiers to a modified bedside PEWS baseline and showed that improvements were witnessed in both accuracy and AUROC on two hidden test datasets.

The results presented showed that similar values were achieved on the 20\% held-out test set from the Desert facility compared to the 100\% held-out test data from the Thunderbird facility. This provides initial evidence for model generalizability, however, further evaluation between facilities at separate locations within the United States, as well as internationally would provide stronger support of model generalization.

While the generalizability of the results are encouraging, a deployed system within a hospital environment that utilized our algorithm would likely still result in a reasonably large number of false positives, based on final threshold selection. In a pediatric general ward setting, a greater number of false positives may be more tolerable than false negatives. However, this choice would need to be at the discretion of the medical care provider and their institution.

There have been previous efforts at using machine learning to construct PICU transfer prediction classifiers. \citep{zhai2014developing} looked at the problem of predicting PICU transfer for newly hospitalized children. In particular, they used a logistic regression classifier to gauge the likelihood of transfer to the PICU in first 24 hours of a child's hospital stay. All inputs into their algorithm (including vital sign information) were assigned into categories, requiring the specification of cut-off boundaries. They evaluated their algorithm within a single hospital setting and achieved strong prediction performance compared to baseline PEWS systems. In this work, we have constructed classifiers that work directly with vital sign measurements as input. Further, we have ensured that our algorithm has been evaluated on retrospective data collected at separate hospital facilities.

There exists a greater body of work for detecting deterioration in adult populations. Churpek  and colleagues have published extensively \citep{churpek2016value,churpek2016multicenter} on the training and evaluation of machine learning algorithms for predicting adverse events in a clinical setting for adult populations. In \citep{churpek2016multicenter}, they compared numerous classifiers to the Modified Early Warning Score (MEWS) \citep{subbe2001validation}. To train their machine learning models they used vital sign measurements as input, as well as lab result values and demographic information (i.e. age). They found that a random forests classifier performed the best, reporting an AUROC of 0.8. The work presented in \citep{churpek2016multicenter} was developed for adult populations (where the average patient age was 60 years old), whereas our algorithm is tailored to pediatric populations where baseline physiological measurements and normal ranges can differ markedly. Further, their work looked at prediction events of transfer to ICU, as well as cardiac arrest and mortality, whereas our algorithm was trained for the purpose of predicting transfer from the general ward to the PICU. 

In a separate work \citep{churpek2016value}, investigated the importance of trend information for predicting deterioration in hospitalized adults. Here, they used only physiologic vital sign information as input to their models and compared the inclusion of trend information such as the change in current value from previous value, mean and standard deviation of previous values and slope information. In general, they found performance improved by including trend information. The work we have presented requires only a single snapshot of physiologic vital sign inputs to make a prediction about PICU transfer. The inclusion of trend information could perhaps further improve our results for pediatric populations, but care needs to be taken to ensure that bias is not introduced into the algorithm stemming from the increased frequency of monitoring that typically occurs when patients are suspected of deteriorating by clinical care staff. 

\subsection{Limitations}

Our work has the following limitations:

\begin{itemize}
\item The number of available instances that result in transfer to the PICU was limited. To aid model training we have balanced our dataset to ensure the number of transfer instances used during model training match the number of non-transfer instances. This inevitably results in discarding non-transfer encounters, which may potentially contain useful information that can further improve model performance. Currently, our models use only a single snapshot of physiologic measurements from a deterioration encounter. It is possible that performance could be improved by including an increased number of snapshots occurring at various time points during the deterioration period.
\item While we have performed an inter-facility retrospective evaluation, our models have only been evaluated on pediatric populations from two hospitals both situated in the United States. Further evaluation is required to ensure the results obtained generalize across pediatric populations at other hospitals, as well as in other countries.
\end{itemize}


\bibliography{mucmd}

\end{document}